# Performance Analysis and Evaluation of Cloud Vision Emotion APIs

Salik Ram Khanal*

Washington State University

Prabin Sharma

University of Massachusetts Boston, Boston, USA, Prabin.sharma001@umb.edu.com

HUGO Fernandes

INESC TEC and University of Trás-os-Montes e Alto Douro Vila Real, Portugal

João Barroso

INESC TEC and University of Trás-os-Montes e Alto Douro Vila Real, Portugal

Vítor Manuel de Jesus Filipe

INESC TEC and University of Trás-os-Montes e Alto Douro Vila Real, Portugal

**Abstract:** Facial expression is a way of communication that can be used to interact with computers or other electronic devices and the recognition of emotion from faces is an emerging practice with application in many fields. There are many cloud-based vision application programming interfaces available that recognize emotion from facial images and video. In this article, the performances of two well-known APIs were compared using a public dataset of 980 images of facial emotions. For these experiments, a client program was developed which iterates over the image set, calls the cloud services, and caches the results of the emotion detection for each image. The performance was evaluated in each class of emotions using prediction accuracy. It has been found that the prediction accuracy for each emotion varies according to the cloud service being used. Similarly, each service provider presents a strong variation of performance according to the class being analyzed, as can be seen with more detail in this articlets.

## 1. Introduction

There are many ways of natural human-to-human communication that can be replicated into human-to-machine communication. Facial expression is one of the most useful and simple ways of communication in our daily life. Facial expression recognition has widespread through research areas like computer vision, machine learning, human-computer interaction, etc. [1]. Computer vision techniques have been implemented for the detection and recognition of objects ([2]). The application of facial expression recognition has extended to fields like neuroscience, psychology, cognition system, sport, etc. From the many facial expressions that express feelings, eight can be categorized as more diverse from each other and as having more diverse meanings [3].

---

In this era of Information and Communication Technology (ICT), many well-known ICT organizations are working on emotion recognition from images and video, including Google and Microsoft Corporation [4]. Many of them offer cloud-based Application Programming Interfaces (API) with support for several programming languages and platforms, allowing the development of apps with simple access to powerful computer vision and machine learning algorithms and resources. The most common architecture is a cloud Representational State Transfer (REST) API that uses Hypertext Transfer Protocol (HTTP) POST operations to request data analysis on images sent in the request. The cloud server replies providing the recognition results, through JavaScript Object Notation (JSON). By calling a REST API and using just a few lines of code, developers can build a client program that is able to recognize emotions, expressed by one or more persons, in the image.

While there are some cloud vision services offering standard online emotion detection, the providers do not reveal enough information about which algorithms they use. In addition, in the literature, there is a lack of studies about the evaluation, and performance comparison between them. In most of the cases, the performance depends on the number of emotion classes, rating scale, face orientation, image quality, etc. making it difficult to compare the providers [5].

In this article, two famous cloud vision APIs, Google Cloud Vision API – Face Detection ([6], [7] ) are analyzed and compared in terms of their emotion prediction performance using the Karolinska Directed Emotional Faces (KDEF) dataset (Lundqvist, Flykt, & Öhman, 1998). Both the Google Cloud Vision API – Face Detection, and the Microsoft Azure - Emotions API are online computer vision APIs that recognize emotions from facial images. The experiments and results presented in this work are only focused on estimating the API's performance considering their accuracy in the classification of emotions from facial expression. The performance is evaluated using metrics such as confusion matrix: sensitivity, specificity, etc. of the prediction of emotion.

## 2.   **Related Work**

One of the pioneer works on facial expression was published by Ekman [3], which defines a set of emotions that are distinct from each other and have different meanings. At first, emotion recognition was done using a tool for measuring facial expression (*facial action coding system, or* FACS), It segments the facial components into small pieces and analyzes each of them. After this first approach, published in 1976 by Ekman and Friesen [8], the same authors published another article about facial movement to and the recognition of facial expressions ([3], [9]). Facial expression recognition has a wide range of applications including communication, medicine, sport, etc. for various purposes. It can also be applied in elderly care with the implementation of autonomous systems ([10]). An important step in emotion recognition is feature extraction. Before extracting the features from the images, it is more important to know the most important features that directly relate to the emotion. Usually, color and texture of the facial images play a major role in the classification of images [11]. Most of the works related to emotion recognition use two main processing steps: feature extraction and classification. In some recent studies, deep learning is being used to recognize the emotions from facial images and video [12].

In ICT business, many companies provide cloud vision services for emotion recognition in real time, whether from facial images or videos. For the creation of client-side applications, these companies provide SDKs for various programming languages. Some famous APIs are Emotients, Affectiva, Emova, Kairos, Microsoft, Google, etc. [4]. Some websites or blogs compare various emotion recognition APIs in terms of the way the APIs are implemented, and the format of the results provided. Casalboni, Keenan, and Virdee-chapman [13-15] compare cloud vision APIs based on their core features and implementation issues. However, there is still a lack of work comparing the classification's performance of the different APIs. There are many web APIs to work in emotion detection. Doerrfeld [4] lists and compares 23 emotion recognition APIs according to their core features. Some services have separate APIs for the recognition of emotions and for face detection [16-18] while some other include emotion recognition services as part of their face detection APIs [19], [20]. Recently, some of them use deep learning techniques for these purposes [21] [18]. Facial feature points are very effective for the classification of emotion and face recognition. Sightcort et al ([22-24]) implemented some machine learning algorithms for the detection of the landmarks in the face. They also provide an SDK for many programming languages, where users must have an authentication key and a free web API for testing purposes.

In the area of facial recognition, the capacity to recognize facial emotion in different poses is a fundamental requirement. Salik et al [25] performed an experiment with a dataset containing 980 images of each type of five poses [full-left, half-left, straight, half-right, and full-right] with seven emotions. They found out the overall recognition accuracy is best performed by Microsoft Azure for straight images. Emotions can be collected using intelligent interface to predict problems related to health or for their safety. Salik [26] implemented a Microsoft Azure - Emotion SDK to detect and recognize the emotions in real time to which can be used for elderly care support. Prabin et al [27] [28] proposed another application of online APIs is on student concentration monitoring.

Luis [29] compared the emotion detection accuracy of OpenCV, Cognitive services and Google Vision APIs. They found out the Open CV implementation got the best performance and can be improved by increasing sample size per emotion during training. Theresa [30] compared and tested three different Facial emotion recognition systems (Azure Face API, Microsoft;

Face++, Megvii Technology; FaceReader, Noldus Information Technology) with human emotion recognition in standardized posed facial expression and non-standardized (emotions extracted from movie scenes). The accuracy of the classification can be estimated using different techniques [32]. The use of confusion matrix is a performance test tool that can be used for the evaluation of the classification process. In this article, confusion matrix is used. The performance of each API is evaluated using parameters like true positives (TP), true negatives (TN), false positives (FP), and false negatives (FN).

## 3. Google Cloud Vision – Face Detection and Microsoft Azure-Emotion API

**Google Cloud Vision API – Face Detection:** Google launched its Application Program Interface (API) for face detection with support for many programming languages, including Java, Python, etc. The API for face detection does not provide a separate request for the emotion recognition, but it can be accessed by some annotation functions included in the face detector. The API classifies the facial expression into four possible emotions (classes): *joy, sorrow, anger, surprise.* It returns the predicted emotion and the corresponding likelihood value expressed as: *very unlikely, unlikely, possible, likely or very likely. Very unlikely* corresponds to a minimum value of confidence for a particular emotion whereas *very likely* is the maximum value of confidence for that emotion [33]. The services can be interactively tested in the web site: https://cloud.google.com/vision/. In the example of Figure 1, the facial expression is classified as *surprise* with a likelihood of very likely, *joy* with likelihood of possible, and so on.

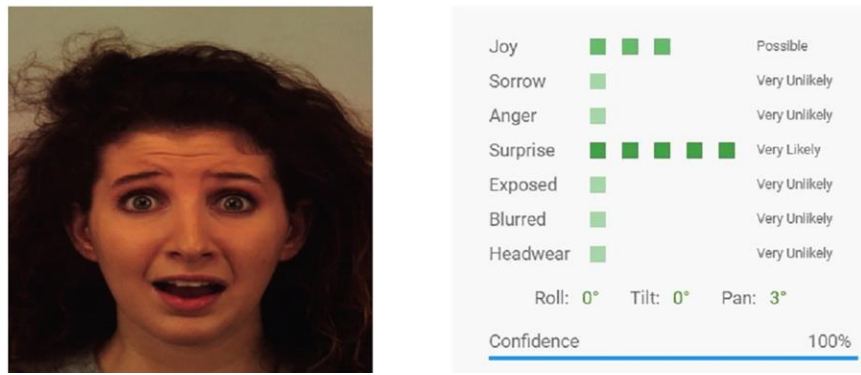

*Figure 1: Response of Google cloud platform – Detecting Faces API free demo version of a sample image*

To access the SDK, special authentication is needed using a key. In our experiment, we used a free authentication key provided by Google, which allows up to a certain number of images during a certain period.

**Microsoft Azure – Emotion API:** Microsoft Corporation lunched an API for emotion detection, which also provides support for many programming languages including Java, Python, C# etc. Images and video from the local machine can be loaded to the cloud server, which returns a list of emotions with the corresponding confidence values for each face detected in the image/video. The confidence (value between 0 and 1) gives the likelihood for each class of emotion. The API classifies the facial expression in eight classes of emotion: *anger, contempt, disgust, fear, happiness, neutral, sadness, and surprise.* In the example of Figure 2, the facial expression is classified as *disgust* with a likelihood of 0.68, *sadness* with likelihood of 0.21, *anger* with likelihood of 0.09, and so on.

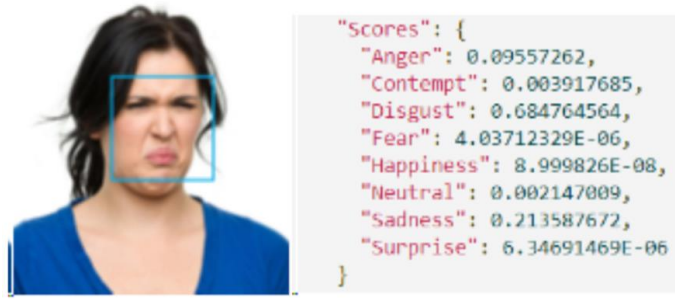

*Figure 2: Response of Microsoft Cognitive Service – Emotion API free demo version of a sample image*

The API gives three options to input images: from a local computer, from a web address and a video file. It can detect more than one faces in one image or video and returns the result for each of the faces [34]. The service can be interactively tested in the web site: https://azure.microsoft.com/en-us/services/cognitive-services/. Microsoft also provides free authentication key up to a certain number of images, which was used in the experiment.

## 4. Methodology

The main objective of this work is to evaluate and compare the performance of two cloud-based APIs for facial emotion recognition. For this purpose, a client program was developed to evaluate the results of emotion classification using images from a public dataset. Figure 3 presents the client program architecture that iterates over the image set, makes the request to the cloud service, caches the result for each image and, finally compiles the global results calculating some metrics.

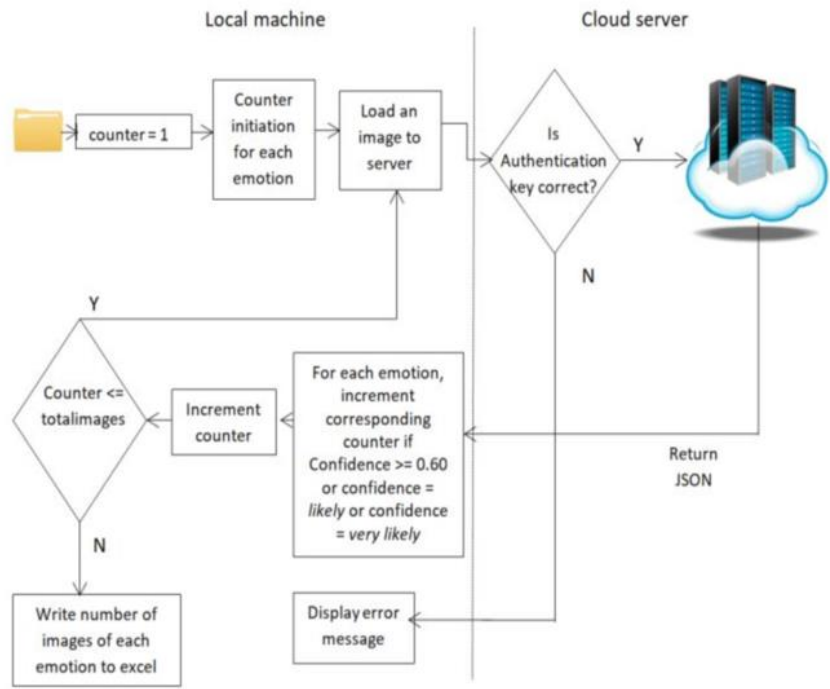

*Figure 3: Client program architecture developed for evaluation of Microsoft Azure - Emotion API and Google Cloud Vision - Face Detection API.*

The software developed, programmatically iterates through all the images of a directory, which contains images with same emotion category, calls API, caches the responses, and save results in an excel file. Therefore, the outputs of the analysis of each sub-folder is stored as an excel table. The format of each excel tab is shown in the next section.

### Dataset Description

Experiments were conducted with a subset of images extracted from the public dataset Karolinska Directed Emotional Faces (KDEF) (Lundqvist, Flykt, & Öhman, 1998), available at http://www.emotionlab.se/resources/kdef. The KDEF dataset has 4900 facial images, which were collected from 35 males and 35 females in two sessions (2 images per emotion). The dataset lumps the facial emotions into seven categories: *neutral, happiness, angry, afraid, disgusted, sadness,* and *surprised* and five camera directions: *full left, half left, straight, half right, full right.* A subset with 980 images of faces only with the frontal plane of face (camera direction: straight) was extracted from the original dataset. As an example of the dataset, Figure 4 shows seven images of the same person with different facial expressions. The person's ID, gender, and the facial emotion of the image are indicated in the filename. The first character indicates the session (A or B); the second character indicates: male (M) or female (F); third and fourth: ID number (01 – 35); fifth and sixth: emotion category - *AF = afraid; AN = angry; DI = disgusted; HA = happy; NE = neutral; SA = sad; SU = surprised;* and the last two characters refers to camera angle in the capture: *FL = full left profile; HL = half left profile; S = straight; HR = half right profile; FR = full right profile.*

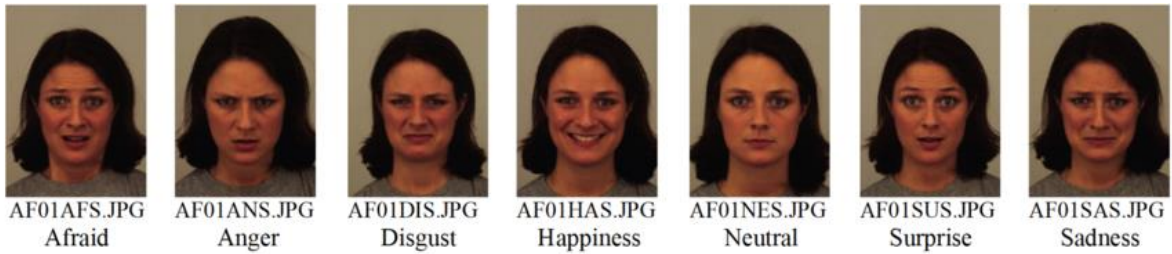

| AF01AFS.JPG | AF01ANS.JPG | AF01DIS.JPG | AF01HAS.JPG | AF01NES.JPG | AF01SUS.JPG | AF01SAS.JPG |
|---|---|---|---|---|---|---|
| Afraid | Anger | Disgust | Happiness | Neutral | Surprise | Sadness |

*Figure 4: Sample facial images of a female with seven emotions with the annotated filename.*

**Experiments**

In the experimental results, the number of correctly predicted images and the number of images that are miss-classified as other emotion etc. are analyzed. One difficulty when comparing the accuracy of prediction is related to the format in which the results are returned by the Google and Microsoft APIs, which are different from each other and from the KDEF dataset. The diversity in number categories, way of returning confidence value and availability of the APIs, and the KDEF dataset is presented in Table 1.

*Table 1: Number of categories and confidence values of Google, Microsoft, and KDEF dataset.*

|  | Google | Microsoft | KDEF Dataset |
|---|---|---|---|
| Number of emotion categories | 4 (*anger, joy, sorrow, surprise*) | 8 (*anger, happiness, sadness, surprise, disgust, neutral, fear and contempt*). | 7 (*anger, happiness, sad, surprise, disgusted, neutral, and afraid*) |
| Confidence value | Five discrete values of confidence: *very unlikely, unlikely, possible, likely or very likely* | Numeric value [0 1] | Not applicable |

Therefore, the outputs of both APIs must be adjusted in order to be comparable to each other. In the case of the Google API, only the emotion with a confidence value of *likely* or *very likely* is considered as a correct prediction. The Microsoft API returns eight classes of emotions, each one with a confidence score (numeric value between 0 and 1) that is converted into five discrete values, similarly to Google, so that the format of the result is the same. Only the result with a confidence value equal or greater than 0.60 is considered as a correct prediction. This adjustment in the APIs outputs has the consequence that images with low confidence value, in all the emotions, are not classified in any of the categories.

## 5.    Results and Discussion

**Google API Evaluation:** To analyze the true positive rate of each emotion, a confusion matrix was drawn for all the emotions that are recognized by the Google API. The set of 980 images contains 140 of each emotion, 70 males, and 70 females. Table 3 shows that *joy* has the best result (100% of correct predictions), while *sadness* and *surprise* have the worst results, respectively with 10% and 24.29% of correct predictions. The confusion matrix for the results obtained from the Google API is shown in Table 2.

*Table 2: Confusion matrix of the prediction obtained from 980 images with the Google API.*

|  |  | Predicted (%) |  |  |  | Not classified |
|---|---|---|---|---|---|---|
|  |  | Anger | Joy | Sorrow | Surprise |  |
| Actual | Anger | **25.71** | 1.43 | 5.00 | 0.00 | 67.86 |
|  | Joy | 0.00 | **100.00** | 0.00 | 0.00 | 0.00 |
|  | Sorrow | 3.57 | 3.57 | **10.00** | 0.00 | 82.86 |
|  | Surprise | 0.00 | 1.43 | 0.00 | **24.29** | 74.28 |

As mentioned in the previous section, the likelihood values of the particular emotion *likely* or *very likely* were considered as correct predictions while other values were considered as not predicted for that emotion. From the confusion matrix drawn in Table 3, 25.27 % of *anger* images were correctly predicted as an *anger*, which is called a true positive (TP) rate. It is observed that 6.43% of *anger* images are incorrectly predicted: 1.43% wrongly classified as *happiness* and 5% wrongly classified as *sorrow*. Since only likelihood values of *likely* or *very likely* were considered as correct predictions, the values in the *Not classified* column represent the percentage of images that have likelihood values below *likely*. From table 3 it can be concluded that 67.84% of *anger* images are not classified as neither of emotion. In case of *joy* or *happiness*, the TP rate is 100% correct. In case of *sorrow*, only 10% of images are correctly predicted, 3.57% are wrongly predicted as *anger* and 3.57% are wrongly predicted as *joy*. From these results, *anger* and *sorrow* emotions are more related to each other and are the most difficult to recognize.

**Microsoft API Evaluation**

The same experiment was done with the same dataset to evaluate performance of the Microsoft Emotion API. Unlike Google, the Microsoft Emotion API classifies facial emotions into eight categories. Since the *contempt* emotion is not in the KDEF dataset, this emotion was not included in the table and, therefore, the table has only the results of 7 emotions. From the 70 input images expressing the *anger* emotion, only 8 images were predicted as *very likely* and 10 as *likely*. Since the classification with the likelihood of *likely* and *very likely* was considered as correct prediction, this means that only 25.7% of *anger* images are correctly classified. Table 3 also shows the miss-classifications of emotion. For example, 10 images with *anger* expression were predicted as *very likely* or *likely* with the *neutral* emotion. It can be concluded that *anger* and *neutral* emotion are related with each other. Similarly, the confusion matrix obtained from the experiment using the Microsoft – Emotion API is drawn in Table 3.

*Table 3: Confusion matrix of prediction obtained in 980 images with Microsoft API*

|        |           | Predicted (%) | | | | | | | |
|--------|-----------|-------|---------|------|-----------|---------|---------|----------|----------------|
|        |           | Anger | Disgust | Fear | Happiness | Neutral | Sadness | Surprise | Not classified |
| Actual | Anger     | **31.43** | 0.00 | 0.00 | 0.00 | 17.86 | 0.00 | 0.00 | 50.71 |
|        | Disgust   | 1.43 | **48.57** | 0.00 | 0.71 | 1.43 | 6.43 | 0.00 | 41.43 |
|        | Fear      | 0.00 | 0.00 | **6.43** | 1.43 | 4.29 | 10.00 | 17.86 | 59.99 |
|        | Happiness | 0.00 | 0.00 | 0.00 | **100.00** | 0.00 | 0.00 | 0.00 | 0.00 |
|        | Neutral   | 0.00 | 0.00 | 0.00 | 0.00 | **100.00** | 0.00 | 0.00 | 0.00 |
|        | Sadness   | 0.00 | 0.00 | 0.00 | 0.00 | 3.57 | **76.43** | 0.00 | 20.00 |
|        | Surprise  | 0.00 | 0.00 | 0.00 | 0.71 | 2.86 | 0.00 | **90.00** | 6.43 |

From Table 3, it is seen that only 31.43% of *anger* images are correctly classified, 17.86% of the images are miss-classified as *neutral* and 50.71% of images are not classified as neither of emotions. Similarly, 48.57 % of *disgust* emotions are correctly classified and 1.43%, 0.71%, 1.43% and 6.43% of the images are miss-classified as *anger*, *happiness*, *neutral*, and *sadness* emotions, respectively, while 41.43 % of the images are not classified as neither of emotions. From these results, it is also seen that neutral and happiness have 100% correct classification while *fear* has the worst result only 6.43% of correct predictions. The results from Table 4 also shows that some emotions are more correlated with each other, which means that there is a higher chance of miss-classification in Microsoft as compared to Google.

It is important to compare the results for male and female to know whether the performance depends on gender or not, so, both APIs were tested with both female and male of 70 images of each, for a similar emotion category as shown in Table 6. Even though Microsoft classifies faces into 8 different emotions, only 4 similar emotions were considered so that the performance analysis of both APIs can be compared. As Google and Microsoft have different names for the same emotion, such as *joy / happiness* and *sorrow / sadness*, the names for these 2 categories were defined as *sadness* and *happiness*, respectively.

From the 70 input images of *anger*, of both male and female, 28.57 % of the images of the female are correctly classified and 22.84% images of males are correctly classified in case of Google Vision API.

*Table 4: True positive rate in male and female with Google Cloud Vision-face detection API and Microsoft Cognitive Services - Emotion API*

|           | Google Vision API Face Detection | | Microsoft Cognitive Services Emotion Detection | |
|-----------|--------|--------|--------|--------|
|           | Female | Male   | Female | Male   |
| Anger     | 28.57  | 22.86  | 25.71  | 37.14  |
| Happiness | 100.00 | 100.00 | 100.00 | 100.00 |
| Sadness   | 8.57   | 11.43  | 84.29  | 68.57  |
| Surprise  | 28.57  | 20.00  | 87.14  | 92.86  |

From Table 6, it is seen that the results of male and female are random and similar, results, so, we can combine the images of both male and female. From Table 4, it is also seen that the classification of *happiness* emotion is easily classified. Except for the *anger* emotion, the result of Microsoft API's is better than the Google's even if it is classified into 8 emotions. From the overall result analysis, the performance of Microsoft Cognitive Service-Emotion API is better than the Google Cloud Vision – Emotion API.

This can also be seen in bar diagram as shown in Figure 3, which provides for a better understanding.

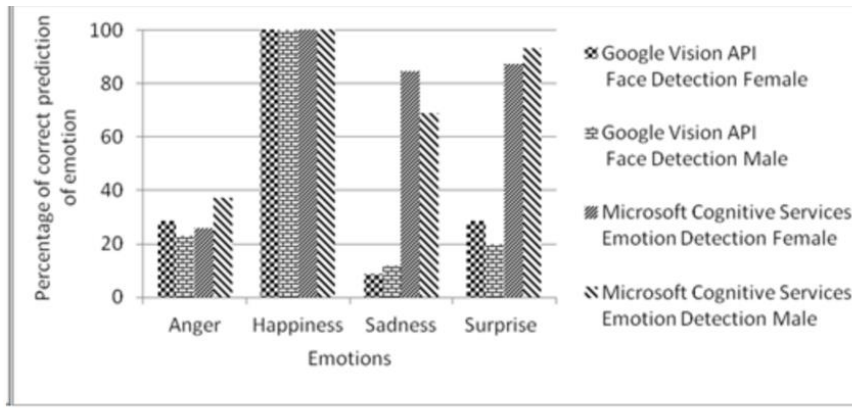

*Figure 5: Bar diagram comparing Google Cloud Vision-Emotion API and Microsoft Cognitive Services -Emotion API, in terms of performance*

Some more parameters from the confusion matrix reflect the realistic analytical result of the prediction. True Positive (TP: percentages of target emotions which are correctly classified), False Negative (FN: percentages of target emotions which are miss-classified as other emotions), False Positive (FP: percentages of other emotions which are miss-classified as target emotion), and True Negative (TN: percentages of other emotion which are classified as other emotion) of both APIs help to analyze the result in more effective way.

*Table 5: FP, FN, TP, and TN of each API, for four emotions*

| | Google | | | | Microsoft | | | |
|---|---|---|---|---|---|---|---|---|
| Emotion | TP | FP | TN | FN | TP | FP | TN | FN |
| Anger | 25.71 | 1.19 | 98.81 | 74.29 | 31.43 | 0.23 | 99.77 | 68.57 |
| Happiness | 100.00 | 2.14 | 97.86 | 0.00 | 100.00 | 0.48 | 99.52 | 0.00 |
| Sadness | 10.00 | 1.67 | 98.33 | 90.00 | 76.43 | 2.73 | 97.27 | 23.57 |
| Surprise | 24.29 | 0.00 | 100.00 | 75.71 | 90.00 | 2.98 | 97.02 | 10.00 |

In case of Google, 25.7% of *anger* emotions are correctly classified as *anger*, 1.19% of other emotions are miss-classified as *anger*, 98.81% of other emotions are classified as another, and 74.29% of anger emotions are miss-classified as another emotion. As these parameters are performance indicators of the classifiers, high values of TP and TN and low values of FP and FN indicate better performance. From Table 5, it is seen that *happiness* has very good performance and *sadness* and *surprise* have some similarities to each other and are more difficult to classify. As compared with Microsoft, in case of *sadness* and *surprise*, the classification performance Google API is worse. Terminologies and derivations from the confusion matrix for each API and for each of emotion provide a more effective indicator of performance. Nine parameters were calculated to test the performance of each API for case (see Table 6).

*Table 6: Calculation of terminologies of confusion matrix for both APIs for four emotions*

| S.N. | Parameters | Formula | Google | | | | Microsoft | | | |
|---|---|---|---|---|---|---|---|---|---|---|
| | | | Anger | Happiness | Sadness | Surprise | Anger | Happiness | Sadness | Surprise |
| 1 | Sensitivity or Recall or Hit rate or True Positive Rate (TPR) | $TPR = \frac{TP}{P} = \frac{TP}{TP+FN}$ | 0.26 | 1.00 | 0.10 | 0.24 | 0.31 | 1.00 | 0.76 | 0.90 |
| 2 | Specificity or True Negative Rate (TNR) | $TNR = \frac{TN}{N} = \frac{TN}{TN+FP}$ | 0.99 | 0.98 | 0.98 | 1.00 | 1.00 | 1.00 | 0.97 | 0.97 |
| 3 | Precision or Positive Prediction Value (PPV) | $PPV = \frac{TP}{TP+FP}$ | 0.96 | 0.98 | 0.86 | 1.00 | 0.99 | 1.00 | 0.97 | 0.97 |
| 4 | Negative Prediction Value (NPV) | $NPV = \frac{TN}{TN+FN}$ | 0.57 | 1.00 | 0.52 | 0.57 | 0.59 | 1.00 | 0.80 | 0.91 |
| 5 | Miss Rate or False Negative Rate (FNR) | $FNR = \frac{FN}{P} = \frac{FN}{FN+TP}$ | 0.74 | 0.00 | 0.90 | 0.76 | 0.69 | 0.00 | 0.24 | 0.10 |
| 6 | Fall Out or False Positive Rate (FPR) | $FPR = \frac{FP}{N} = \frac{FP}{FP+TN}$ | 0.01 | 0.02 | 0.02 | 0.00 | 0.00 | 0.00 | 0.03 | 0.03 |
| 7 | False Discovery Rate (FDR) | $FDR = \frac{FP}{FP+TP} = 1\text{-}PPV$ | 0.04 | 0.02 | 0.14 | 0.00 | 0.01 | 0.00 | 0.03 | 0.03 |
| 8 | False Omission Rate (FOR) | $FOR = \frac{FN}{FN+TN} = 1\text{-}NPV$ | 0.43 | 0.00 | 0.48 | 0.43 | 0.41 | 0.00 | 0.20 | 0.09 |
| 9 | Accuracy (ACC) | $ACC = \frac{TP+TN}{TP+FN+TN+FP}$ | 0.62 | 0.99 | 0.54 | 0.62 | 0.66 | 1.00 | 0.87 | 0.94 |

The true positive rate (TPR) of Google is higher than Microsoft's. TPR is the positive index of performance. For example, the TPR of *happiness* in both APIs is 1, that is, 100% correct classification. All the *happiness* emotions were classified only as happiness. The true negative rate (TNR) of all the emotions in both APIs seems to be similar, close to 1. It indicates that each other emotions are predicted, as they should.

## 6.  Conclusion and future work

In case of Google API, *joy* or *happiness* has 100% true positive rate and *joy* and *surprise* are less related with *anger*. The true positive rate of *sorrow* is worst as compared with other emotions. In case of Microsoft API, *anger* and *happiness* is less similar to each other whereas *anger* and *surprise* are more similar to each other. The true positive rate of both emotions *happiness* and *neutral* is 100% whereas *fear* has least true positive rate. The result of male and female is not significantly different in each emotion in case of both APIs. From the results of these experiments, it may be concluded that the Microsoft Cognitive Services - Emotion API provides better results than Google Cloud Vision - Emotion API, in most of the cases, even if it classifies into 8 emotions. *Happiness* is correctly predicted in every case, but *anger* and *sadness* are more similar to each other and difficult to classify. Therefore, there are many miss-classifications on *anger* and *sadness* emotions. Particularly, in the case of *anger*, there are many miss-classifications. In future work, the evaluation can be extended to test results of both male and female with video and other forms of emotional information input. As there are more than 20 web APIs which are working with emotion recognition, this work can be extended to analyze more emotion APIs in terms of prediction accuracy as well as other parameters.

**Acknowledgements**